\documentclass{article}

\usepackage{arxiv}

\usepackage[utf8]{inputenc} 
\usepackage[T1]{fontenc}    
\usepackage{hyperref}       
\usepackage{url}            
\usepackage{booktabs}       
\usepackage{amsfonts}       
\usepackage{nicefrac}       
\usepackage{microtype}      
\usepackage{lipsum}
\usepackage{graphicx}
\graphicspath{ {./images/} }
\usepackage{algorithm}
\usepackage{algorithmic}
\usepackage{subcaption}
\usepackage{siunitx}

\title{Human / AI interaction loop training:
 New approach for interactive reinforcement learning
}

\author{
 Neda Navidi \\
  AI Redefined, AI-R Inc.\\
  400 McGill st., Montreal, Canada \\
  \texttt{neda@ai-r.com} \\
}

\begin{document}
\maketitle
\begin{abstract}
Reinforcement-Learning (RL) in various decision-making tasks of Machine-Learning (ML) provides effective results with an agent learning from a stand-alone reward function. However, it presents unique challenges with large amounts of environment states and action spaces, as well as in the determination of rewards. This complexity, coming from high dimensionality and continuousness of the environments considered herein, calls for a large number of learning trials to learn about the environment through RL. Imitation-Learning (IL) offers a promising solution for those challenges using a teacher. In IL, the learning process can take advantage of human-sourced assistance and/or control over the agent and environment. A human teacher and an agent learner are considered in this study. The teacher takes part in the agent’s training towards dealing with the environment, tackling a specific objective, and achieving a predefined goal. Within that paradigm, however, existing IL approaches have the drawback of expecting extensive demonstration information in long-horizon problems. This paper proposes a novel approach combining IL with different types of RL methods, namely State–action–reward–state–action (SARSA) and Asynchronous Advantage Actor-Critic Agents (A3C), to overcome the problems of both stand-alone systems. It is addressed how to effectively leverage the teacher’s feedback – be it direct binary or indirect detailed – for the agent learner to learn sequential decision-making policies. The results of this study on various OpenAI-Gym environments show that this algorithmic method can be incorporated with different combinations, significantly decreases both human endeavor and tedious exploration process. 
\end{abstract}


\section{Introduction}
\paragraph{}
Reinforcement Learning (RL) in various decision- making tasks, provides effective and powerful results with learning an agent from stand-alone reward function. However, it suffers from large amount of environment state, action space, high  and implicity of rewards for real complex environments. The complexity, which is retrieved from high dimensionality and continuousness of real environment,  causes RL’s requirement to the large number of learning trials to understand and learn the environment \cite{mnih2016asynchronous}. A promising solution for the limitation is addressed by Imitation Learning (IL) and exploiting teacher feedback. In IL, the learning process can take advantages of human assistance and control over the agent and the environment. In this study, human  considered as a teacher who teach a learner to deal with environment to tackle a specific object.
\paragraph{}
A teacher can express his feedback to improve the policy in two main methods, namely direct dual feedback and indirect detailed feedback. While in the first method teacher evaluates the agent actions by sending back rewards (positive or negative), in the second method, he can demonstrate the way to complete a task to the agent by actions \cite{mericcli2010complementary,christiano2017deep}. One of main limitations of existing IL approaches is that they may expect extensive demonstration information in long-horizon problems.Our proposed approach leverages integrated RL-IL structure (see Fig. 1) to overcome the RL and IL limitations simultaneously. Also, the approach considers both cases where the agent does or does not need human feedback. Our key design principle is a cooperative structure, in which feedback from the teacher is used to improve the learner’s behavior, improve the sample efficiency and speed up the learning process by IL-RL integration   (See Fig. 2).

\begin{figure} 
    \centering
    \includegraphics[width=150mm]{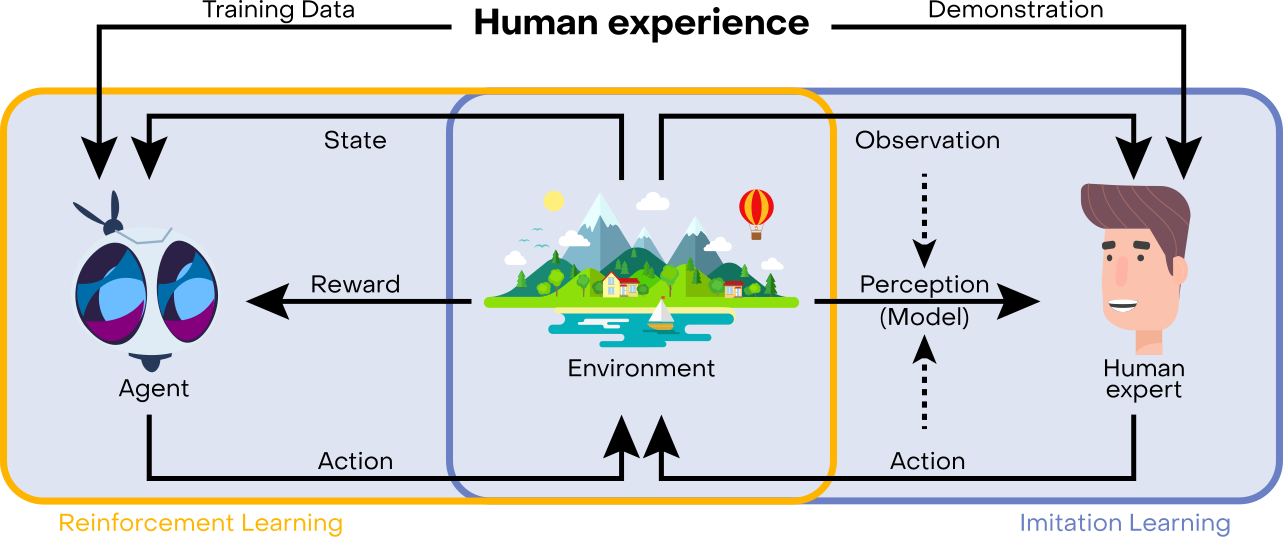}
    \caption{ Reinforcement Learning-Imitation Learning (RL-IL) integration structure}
\end{figure}
\paragraph{}
The teacher assistance considers both direct dual feedback, with positive and negative reward, and indirect detailed feedback, with access to action domain feedback using online policy IL process. Management of teacher’s feedback in the “feedback management” block is one of the features of the structure (see Fig. 2). Also this structure reflects the online teacher feedback as soon as the learner takes an action and deals with quantity of teacher’s feedback. 
\paragraph{}
This paper begins by overviewing the related work on RL and IL in section 2. It is continued by formalizing the problem of imitation learning and details of the proposed structure (Section 3). The proposed structure is validated and compared with RL stand-alone in section 4. The experimental validation and analysis the results are presented in section 5 and 6.
\section{Related Works}
\label{sec:headings}
\paragraph{}
 Having an agent learn behavior from a standalone reward, which is the main concept of Reinforcement Learning (RL), is particularly difficult in a complicated environment. It mainly suffers from high dimensionality of environment spaces in challenging tasks. Also, the definition of reward function in real-word applications is very complicated and implicit. Contribution of human and agents in the form of using the human knowledge in the training loop by Imitation Learning (IL) is a promising solution to improve the data efficiency and to gain a robust policy \cite{ross2011reduction}.
\paragraph{}
In IL, the agent observes, trusts and replicates the teacher’s behavior. A typical method of IL, which is presented as Behavioural Cloning (BC) or Learning from Demonstration (LfD), the goal is to train a classifier-based policy to predict the teacher's actions. In BC, features are list of environment’s observations and the labeled-data are actions performed by the teacher. However, the statistical learning assumption affected by ignoring the relationship of current action and next states during execution of the learned policy, causes poor performance of this method \cite{moore2019behavioural,englert2013model}.
\paragraph{}
Forward training algorithm in IL has been introduced to train one policy at each time step to achieve a non-stationary policy. In the training algorithm, the agent learns how to imitate the teacher behaviour in the states generated by the previous learned policies \cite{ross2010efficient}. The main disadvantage of the forward training algorithm is that it requires investigate the environment over all periods, regardless of the horizon size. In fact, considering the non-stationary policy in this model causes its impracticality in real-world applications with large horizon.
\paragraph{}
Search-based Structured Prediction (SEARN) learns a classifier to choose the search optimally. This model outperforms the traditional search-based methods which first learn some sort of global model and then start exploring. SEARN follows the teacher's action at the beginning of a task. Then it aggregates more demonstrative data iteratively to obtain an updated policy. It generates new episodes to create a combination of previous models and the teacher behaviour \cite{daume2009search}.  However, the optimistic prediction, as a result of the difference between its initial value and the optimal policy, is the main drawback of this learning method.
\paragraph{}
Stochastic Mixing Iterative Learning (SMILe) has been proposed to improve the forward training algorithm using SEARN’s benefits with a straightforward implementation and less dependency to interact with a teacher. After several iterations, the method utilizes a geometric policy by training a stochastic policy \cite{ross2010efficient}. While the training process can be interrupted at any time, it suffers from instability of the model because of the stochastic policy assumption.
\paragraph{}
Two popular IL algorithms called Data-set Aggregation (DAGGER) and Generative Adversarial Imitation Learning (GAIL) \cite{ross2011reduction,ho2016generative} introduce new approaches for incorporating teacher experience. These papers have proposed iterative algorithms for online learning approach to train a stationary deterministic policy. It has been proved that the combination of the algorithms with reduction-based approaches outperforms the policy findings in sequential settings, thanks to reusing existing supervised learning algorithms. 
\paragraph{}
DAGGER performs well for both complex and simple problems, however, the information may not be intuitive from the states \cite{attia2018global}. Also, GAIL has presented considerable achievement on imitation learning especially for complex continuous environments. But it suffers from the requirement of a huge amount of interactions during the training. Furthermore, it is very time-consuming in real-world applications which it is needed more interactions between agent and environment to achieve an appropriate model \cite{sasaki2018sample}
\paragraph{}
\cite{judah2012active} represents that Reductions-based Active Imitation Learning (RAIL) consists of N iterations which each iteration has a specific stationary policy over time steps and has a significant difference with previous iteration. This method provides a small error at the expert actions prediction considering the state distribution of the former iteration. Nevertheless, the results in \cite{judah2012active} can be faulty and impractical in some cases due to the unlabeled state distributions in the previous iterations.
\paragraph{}
As it has been presented in the research studies above, all IL methods mostly suffer from instability of the model because of the stochastic policy assumptions. Also, the labeled-information needs lead to necessity of expert human to annotate the data-set.
These two main drawbacks prevents using IL for high-dimensional high-frequency real-world applications. Fortunately, a promising solution is integration of IL with RL to overcome these aforementioned limitations.
\paragraph{}
The idea of exploiting IL to increase the speed of convergence of RL has been considered in \cite{hester2018deep}. However, it considers the stochastic  policy and uses IL as a “pre-training” solution to speed up the convergence .Considering IL as a pre-training step have been conducted on reward reshaping \cite{brys2015reinforcement}, policy reshaping \cite{griffith2013policy}and knowledge transfer \cite{wang2018transferring} with teacher feedback. 
\paragraph{}
\cite{ng1999policy} uses a reward shaping method which is one of the significant aspects of RL. \cite{pilarski2011online} describes that this method is an accepted way for human feedback in RL, but it causes some issues that human feedback signals may contain inconsistency, infrequency, ambiguity and insufficiency \cite{ng1999policy}. As an example, translating statements into rewards may be difficult and unclear; accordingly \cite{isbell2006cobot} tries to solve this problem considering a drift parameter to reduce the dependency on human feedback signals. To overcome some of the aforementioned limitations, \cite{thomaz2008teachable} proposed an UNDO function as a policy feedback which contains a reversible feedback signal for agents. The results in \cite{pilarski2011online} presents that the human feedback signals can improve RL algorithms by applying them in the process of action selection. Some recent references uses the human feedback as an optimal policy instead of a reward shaping method like agent’s exploration seed \cite{taylor2011integrating}, inverse RL control \cite{ziebart2008maximum}, and even as a substitute of exploration \cite{wang2018transferring,ng1999policy}.
\paragraph{}
The core of the paper provides an accessible and effective structure for the agent to get expert with teacher help and advice. It also addresses a set of generic questions, namely what should be imitated, how the agent may imitate, when is the time of imitating, and who is trustworthy to imitate. Also, when teachers are available; this paper addresses how their feedback can be the most effectively leveraged.
\section{Teacher Assistance Structure}
\label{sec:headings}
\paragraph{}
The proposed structure exploits teacher feedback as a rectification of the action domain of the learner; as soon as an action was performed by the agent, this teacher feedback improves the online policy. It can also infer the policy of the agent from infrequent teacher feedback. It is considered and formulated four main characteristics of human teacher feedback and their related effects; namely, duality, reaction time delay, contingency and instability.
\begin{figure} 
    \centering
    \includegraphics[width=150mm]{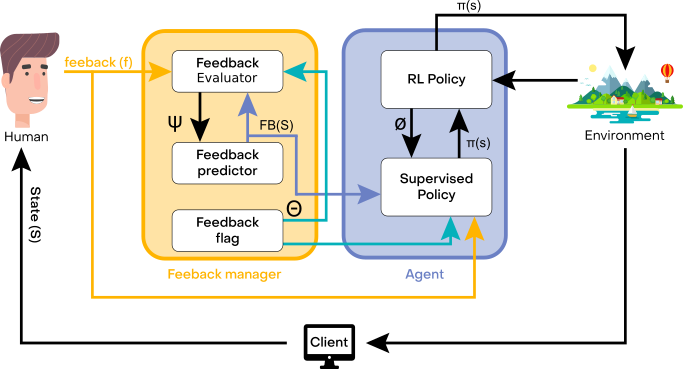}
    \caption{ Proposed structure for human/AI interaction}
\end{figure}
\paragraph{}
Although several studies consider a range for teacher feedback, like {very bad, bad, good, very good} or {-100,-50, 50, 100}, giving feedback by humans from a range is very complicated and requires very good knowledge of the task and environment. This study takes advantage of duality feedback when a human teacher is satisfied with the taken decision by agent or not. This kind of feedback can be sent by expert or non-expert human in order to its simplicity of transferring the knowledge. 
\paragraph{}
The next feature which is considered, is the reaction time delay of human to send feedback. Several studies like \cite{chevalier2018babyai}presents the sample-efficiency for training neural reinforcement when it is pre-trained by an expert using supervised learning method. In fact, in different IL algorithms like DAGGER and GAIL which are based on offline learning, there is no need to consider the reaction time delay of human in the model. In the mentioned algorithms just an expert should prepare a time-consuming metadata before starting the training process. Also, the unprofessional feedback comes from an inexpert teacher can ruin the training process.
\paragraph{}
The proposed structure for human/AI interaction presents a methodology to interact AI agent with human (teacher expert or not) completely online, we should deal with the delay in reaction of humans \cite{cauchard2016emotion,jaderberg2019human}. Using the teacher feedback in online training without recognizing that delay can make the training process impractical. However the reaction time delay is not constant and it would vary depending on teacher personality, teacher knowledge, complexity of the environment, and ambiguity of actions.
\paragraph{}
In addition to the  reaction time delay, it is dealt with contingency of the human teacher feedback as a feature of reactive manner. In order to  limited human patience, mostly the human teacher stops to send positive feedback while the agent takes actions correctly. Also the frequency of releasing the feedback can vary based on the human preference \cite{peng2016need}. So the proposed methodology considers a module named “feedback predictor”  See Fig. 2) to present the contingency and stochasticity of correct feedback which is sent in a specific timestamp.
\paragraph{}
The details of the structure is explained as follows (See Fig. 2):
\begin{itemize}
\item Feedback predictor: This section gathers the previous feedback and advice, and predict the next action to be taken by the human.
\item Supervised policy: This module can improve the variables of the RL policy;
\item Feedback evaluator: This module can upgrade the variables of the supervised policy module;
\item Feedback flag: This module can manage and control the time-lags and postponement of human respond;
\item RL policy: It considers an on-policy algorithm of Q-learning and one policy-based RL algorithm.
\end{itemize}
\paragraph{}
As soon as the agent picks an action, the supervisor can observe the outcome of that action on the environment and send his feedback. This feedback ($f$) is a positive, negative or neutral  value to present that the last performed action should be modified increasingly or decreasingly. The neutral value is considered when the teacher is not available or he prefers not to expose his idea. The environment state $(S)$, performed action, $A(S)$, and teacher feedback $(f)$ are sent back to “feedback evaluator” and “supervised policy” sections to update the $\Phi$   \textrm{and}   $\Psi$.
\subsection{Feedback Predictor}
\paragraph{}
$FB(s)$ shows the policy learned in “feedback manager” box and is able to predict the next feedback of the teachers by observing the current state and agent action. As it is considered the dual teachers feedback in range [-1, 1],  -1 shows that the teachers are not satisfied with the action taken by the agent, so they send a request to the agent to stop or reduce it. On the other hand, they send back +1 whenever the taken action is convincing, so the teachers encourage the agent to continue it.Also, an adjustable learning rate is consideredto improve the online and offline models on the online training data-set is monitored by the learning algorithm and the learning rate can be adjusted in response. The policy is formulated by equation (1):
\begin{equation}
 FB(O,A) =\psi ^{T}\theta (O,A)
\end{equation}
where, $FB(O,A)$,  $\psi$ and  $\theta (O,A)$ are the teacher feedback policy, the parameters vector and the probability density function delay of the human's feedback signal, respectively. Details about these parameters are explained in the next sections.
\subsection{Supervised Policy}
\paragraph{}
The policy can be updated and modified directly by supervised policy based on supervised learning methods. In fact, the agent can change its actions based on state-action training pairs provided real-time by the supervisor; without considering the value of those training data. This element can improve the model parameters, using State–action reward–state–action (SARSA) from value-based RL algorithms or Asynchronous Actor-Critic Advantage Agent (A3C) from policy-based algorithms.
\paragraph{}
RL algorithms are required for the optimization process, whereas the teacher helps the agent to gain a level of skill while the RL algorithms have poor estimation of value functions. The supervised policy module provides both error information for the agent as long as the actions for the environment.
\paragraph{}
The agent receives evaluations of its behaviours that can help to perform the given task. Whenever the agent gets professional on the task, the teacher gradually withdraws the additional feedback to shape the policy toward optimality for the true task. The error of the prediction is not clear because of the uncertainty of the quantified human feedback. It is considered in Equation (2):
\begin{equation}
    e(t) = r(t)*k
\end{equation}
Where, the $e(t)$,  $r(t)$ and $k$ present the  prediction error, error sign extracted from human feedback and constant error value predefined by the user, respectively.
\subsection{ Feedback Evaluator}
The responsibility of feedback evaluator is to update the parameters vector ($\psi$) of the teacher feedback policy  ($FB(s)$). $FB(s)$ can be calculated by multiplying the teacher feedback and the parameters vector. It can be rewritten Equation (3):
\begin{equation}
    \Delta \psi = \gamma (f-FB(s))\cdot (\frac{\delta (FB(s))}{\delta \psi})
\end{equation}
where, $\gamma$ is the adjustable learning rate and can be observed by:
\begin{equation}
    \gamma(s)= \left \| FB(s) \right \|+b
\end{equation}
where, $b$ shows the predefined value of learning rate as the bias of the model. The variation between the actual teacher feedback and the predicted teacher feedback (calculated from the teacher feedback policy) is considered as prediction error.
\subsection{Feedback Flag}
\paragraph{}
The action space of control system generally are dichotomized in continuous and discrete. Continuous control systems mostly are designed to deal with continuous action space, specially in high-frequency environments like video games. Speed of system, time-lay of call-response, and non-constancy of human-response rate make communication of the system and human very difficult in these environments. To deal with this problem, The “Feedback flag” module is presented to bufferize and integralize the several past state-action tuples. Each past state-action tuple is weighted with the corresponding probability that characterizes the delay in teacher response and it is shown by $RD(t)$.
\paragraph{}
It is used by the “Feedback evaluator” and the “Supervised policy”. The teacher feedback function is a linear model (Equation 1 and Fig. 2). The uncertainty of feedback’s receiving-time , time $t$ is defined as $t_{1}<t<t_{n}$ , and directly affects the agent that is trying to attach the reward signal to a specific action in time. This feedback could in fact be attached to any prior action at time $t-1, t-2, ..., t-n$. This is why we need to use ($\Theta$) to define the delay of the human's response signal:
\begin{equation}
    \Theta _{t_{0}}= \int_{t-t_{0}}^{t_{0}}RD(t)
\end{equation}
where, $\Theta$ is the density of the continuous human's response. 
\subsection{Reinforcement Learning (RL) Policy}
\paragraph{}
In RL, we consider a predefined environment. In that environment, an agent takes actions and reactions sequentially to complete a task, using its observations of the environment and the rewards gets from it. The agent can choose an action from action space, $A_{s}= {A_{1} , A_{2}, ..., A_{n} }$. That action passes to the stochastic environment and return a new observation space, $O_{s}= {O_{1} , O_{2}, ..., O_{n} }$ and reward,  $R_{s}= {R_{1} , R_{2}, ..., R_{n} }$ . At each step, the agent observes the current state game state and cannot understand the whole task by observing just the current game state. Also, Marcov Decision Process (MDP) is the fundamental of RL. MDP can be used in a cooperative structure for the decision-making tasks to partly or completely control and balance agents.
\paragraph{}
The first RL algorithm in this study is SARSA and its details are presented in Algorithm 1. SARSA is very similar to Q-learning. The key difference between them is that an on-policy whereas a Q-learning is a class of off-policy Temporal Difference (TD). It implies that SARSA learning process is dealing with the actions taken by the current policy instead of the greedy policy. So the SARSA update (see Equation. 6) does not consider the maximum value and greedy policy (Equation 2).
\begin{equation}
    Q(O_{n},A_{s}) = Q(O_{n},A_s)+\alpha[R_{n+1}+ \gamma Q(O_{n+1},A)-Q(O_{n}, A_{s})]
\end{equation}
On the other hand, Q-learning consists of a multi-layer neural network (NN) which its inputs would be states of an environment and the outputs would be the action value, $Q(s,\theta)$ , which $\theta$ is the parameters of $NN$. In fact, Q-learning updates for the parameter after taking action $A_{n}$ after observing the state, $O_{n}$, and receive an immediate reward  $R_{n}$ and $Q_{n+1}$. So the update equation is given by:
\begin{equation}
    Q(O_{n},A_{s}) = Q(O_{n},A_s)+\alpha[R_{n}+\gamma.max Q(O_{n+1},A)-Q(O_{n}, A_{s})]
\end{equation}
This equation shows that the policy considered to select an action is a greedy policy calculated by Equation (8):
\begin{equation}
    max Q(O_{n+1},A)-Q(O_{n}, A_{s})
\end{equation}
\paragraph{}
The second RL algorithm in this study is Asynchronous Advantage Actor-Critic Agents (A3C) as a policy gradient algorithm with a special focus on parallel training. In A3C, the critic part learns the value function while the actor part is trained in parallel and become synchronized with the global parameters sequentially. In A3C, there is a loss function for state value to minimize the Mean Square Error (MSE) (Equation 9) and it is the baseline in the policy gradient update. Finally, the gradient descent can be applied to find the optimal value. For more details about the A3C see the Algorithm 2.
\begin{equation}
    J_{V}(\omega)= (G_{t}-V_{\omega}(s))^2
\end{equation}

\begin{algorithm}
\caption{State–action reward–state–action (SARSA)}
\begin{algorithmic}
\label{Algorithm SARSA}
\REQUIRE 
\STATE Observations: $O_{s}= {O_{1} , O_{2}, ..., O_{n}}$,
\STATE Actions: $A_{s}= {A_{1} , A_{2}, ..., A_{n} }$, 
\STATE Reward Function: $R_{s}= {R_{1} , R_{2}, ..., R_{n} }$,
\STATE Transition: $T_{s}= {T_{1} , T_{2}, ..., T_{n} }$,
\STATE Initialization:
\STATE Learning rate: $\alpha \in [0,1]$, initialized with = $0.1$
\STATE Discount factor: $\gamma \in [0,1]$
\STATE Balancing rate:  $\lambda \in [0,1]$; Trade off between Temporal-Difference and Monte-Carlo Process Q-learning $(O, A, R, T, \alpha, \gamma, \lambda)$
\WHILE{ $Q$ is not converged}
\STATE select $[O, A] \in O_{s}$
\WHILE{ $O$ is not terminated}
\STATE $R \leftarrow R(O,A)$
\STATE $O' \leftarrow T(O,A)$
\STATE $A' \leftarrow \phi(O')$
\STATE $e(O,A) \leftarrow e(O,A)+1$
\STATE $\sigma \leftarrow \lambda Q(O_{n+1},A_{n+1})- Q(O_{n},A_{n})$
\FOR{$[O', A'] \in O_{s}$}
\STATE $ Q(O',A') \leftarrow Q(O',A')+ \alpha * \gamma * \sigma * e(O',A') $
\STATE $ e(O',A') \leftarrow \gamma* \lambda*e(O',A')$
\STATE $ O' \leftarrow O$
\STATE $A' \leftarrow A $
\ENDFOR
\ENDWHILE
\ENDWHILE
\RETURN $Q$
\end{algorithmic}
\end{algorithm}

\begin{algorithm}
\caption{Asynchronous Advantage Actor-Critic Agents (A3C)}
\begin{algorithmic}
\label{Algorithm A3C}
\REQUIRE 
\STATE requirements of algorithm \ref{Algorithm SARSA},
\STATE Initialization:
\STATE Meta parameters: $\theta$, $\omega$
\WHILE{ $T< T_{max}$}
\STATE $\theta \leftarrow 0$ and $\omega \leftarrow 0$
\WHILE{ $O$ is not terminated}
\STATE $A' \leftarrow \phi(O')$
\STATE $R \leftarrow R(O,A)$
\STATE $O' \leftarrow T(O,A)$
\FOR{$i =t-1: t_{start}$}
\STATE update $R, d\theta and d\omega$
\ENDFOR
\ENDWHILE
\ENDWHILE
\RETURN $Q$
\end{algorithmic}
\end{algorithm}

\begin{algorithm}
\caption{Hybrid State–action reward–state–action imitation learning (SARSA/IL)}
\begin{algorithmic}
\label{Algorithm Hybrid SARSAIL}
\REQUIRE 
\STATE  Exact requirements of algorithm \ref{Algorithm SARSA},
\STATE Initialization:
\STATE Same as algorithm 1,
\WHILE{ $Q$ is not converged}
\STATE select $[O, A] \in O_{s}$
\WHILE{ $O$ is not terminated}
\STATE $R \leftarrow R(O,A)$
\STATE $O' \leftarrow T(O,A)$
\STATE $A' \leftarrow \phi(O')$
\STATE $e(O,A) \leftarrow e(O,A)+1$
\STATE $\sigma \leftarrow \lambda Q(O_{n+1},A_{n+1})- Q(O_{n},A_{n})$
\FOR{$[O', A'] \in O_{s}$}
\IF{$f$ is exist:}
\STATE Consider Equation (5)
\STATE Consider Equation (1)
\STATE Consider Equation (3)
\STATE Consider Equation (4)
\ENDIF
\STATE $ Q(O',A') \leftarrow Q(O',A')+ \alpha * \gamma * \sigma * e(O',A') $
\STATE $ e(O',A') \leftarrow \gamma* \lambda*e(O',A')$
\STATE $ O' \leftarrow O$
\STATE $A' \leftarrow A $
\ENDFOR
\ENDWHILE
\ENDWHILE
\RETURN $Q$
\end{algorithmic}
\end{algorithm}

\begin{algorithm}
\caption{Hybrid Asynchronous Advantage Actor-Critic Agents Imitation Learning (Hybrid A3C/IL)}
\begin{algorithmic}
\label{Algorithm HybridA3CIL}
\REQUIRE 
\STATE requirements of algorithm \ref{Algorithm SARSA},
\STATE Initialization:
\STATE Meta parameters: $\theta$, $\omega$
\WHILE{ $T< T_{max}$}
\STATE $\theta \leftarrow 0$ and $\omega \leftarrow 0$
\WHILE{ $O$ is not terminated}
\STATE $A' \leftarrow \phi(O')$
\IF{$f$ is exist:}
\STATE Consider Equation (5)
\STATE Consider Equation (1)
\STATE Consider Equation (3)
\STATE Consider Equation (4)
\ENDIF
\STATE $R \leftarrow R(O,A)$
\STATE $O' \leftarrow T(O,A)$
\FOR{$i =t-1: t_{start}$}
\STATE update $R, d\theta and d\omega$
\ENDFOR
\ENDWHILE
\ENDWHILE
\RETURN $Q$
\end{algorithmic}
\end{algorithm}

\section{Experiments and Results}
\begin{figure}
     \centering
     \begin{subfigure}[b]{0.49\textwidth}
         \centering
         \includegraphics[width=\textwidth]{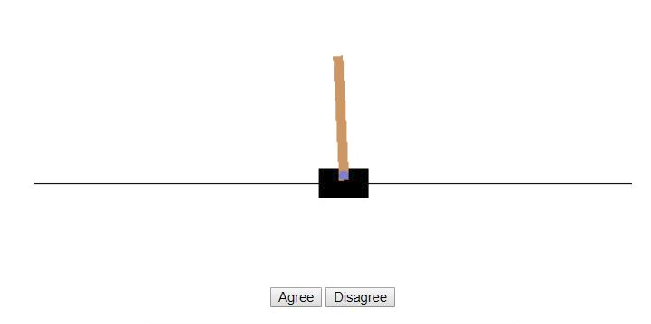}
         \caption{$ $}
        \label{fig:a}
     \end{subfigure}
     \begin{subfigure}[b]{0.49\textwidth}
         \centering
         \includegraphics[width=\textwidth]{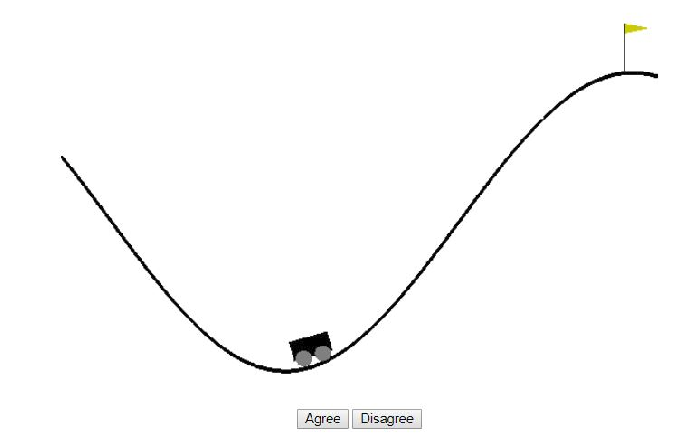}
         \caption{$ $}
        \label{fig:b}
     \end{subfigure}
 \centering
 \caption{Continuous classic OpenAI-Gym environment: (a) Continuous Cart-pole (b) Continuous Mountain Car in the proposed framework}
\end{figure}

The performance of the proposed algorithms \ref{Algorithm Hybrid SARSAIL} and \ref{Algorithm HybridA3CIL} is evaluated on two separate use-cases: 
\begin{itemize}
\item Continuous classic cart-pole OpenAI-Gym environment 
\item Continuous classic mountain-car OpenAI-Gym environment. 
\end{itemize}
\paragraph{}
Continuous cart-pole and mountain-car are used in this study as continuous classical OpenAI-Gym environment. The objective of the cart-pole system is to adjustablely control the cart by taking continuous and unlimited actions. The cart has two degrees of freedom (DoF) to balance with the horizontal axes. The system state is parameterized by orientation, position and velocity of both pole and card.The stability of this system is defined by orientation from \ang{-12} to \ang{12}, and position deviation between $-2.4$ to $2.4$  (See Fig \ref{fig:a}). Whenever the system gets unbalanced, a negative signal as a punishment acknowledge sent back to the system and it will be reset. 
\paragraph{} 
The next OpenAI-Gym considered in this study is continuous Mountain-car illustrated in Fig \ref{fig:b}. This environment presents a car on a sinuous curve track, located between two mountains. The goal is to drive up the right mountain; however, the car's engine is not strong enough to scale the mountain in a single pass. Therefore, the only way to succeed is to drive back and forth to build up momentum. Here, the reward is greater if you spend less energy to reach the goal.

\begin{figure}
     \centering
     \begin{subfigure}[b]{0.49\textwidth}
         \centering
         \includegraphics[width=\textwidth]{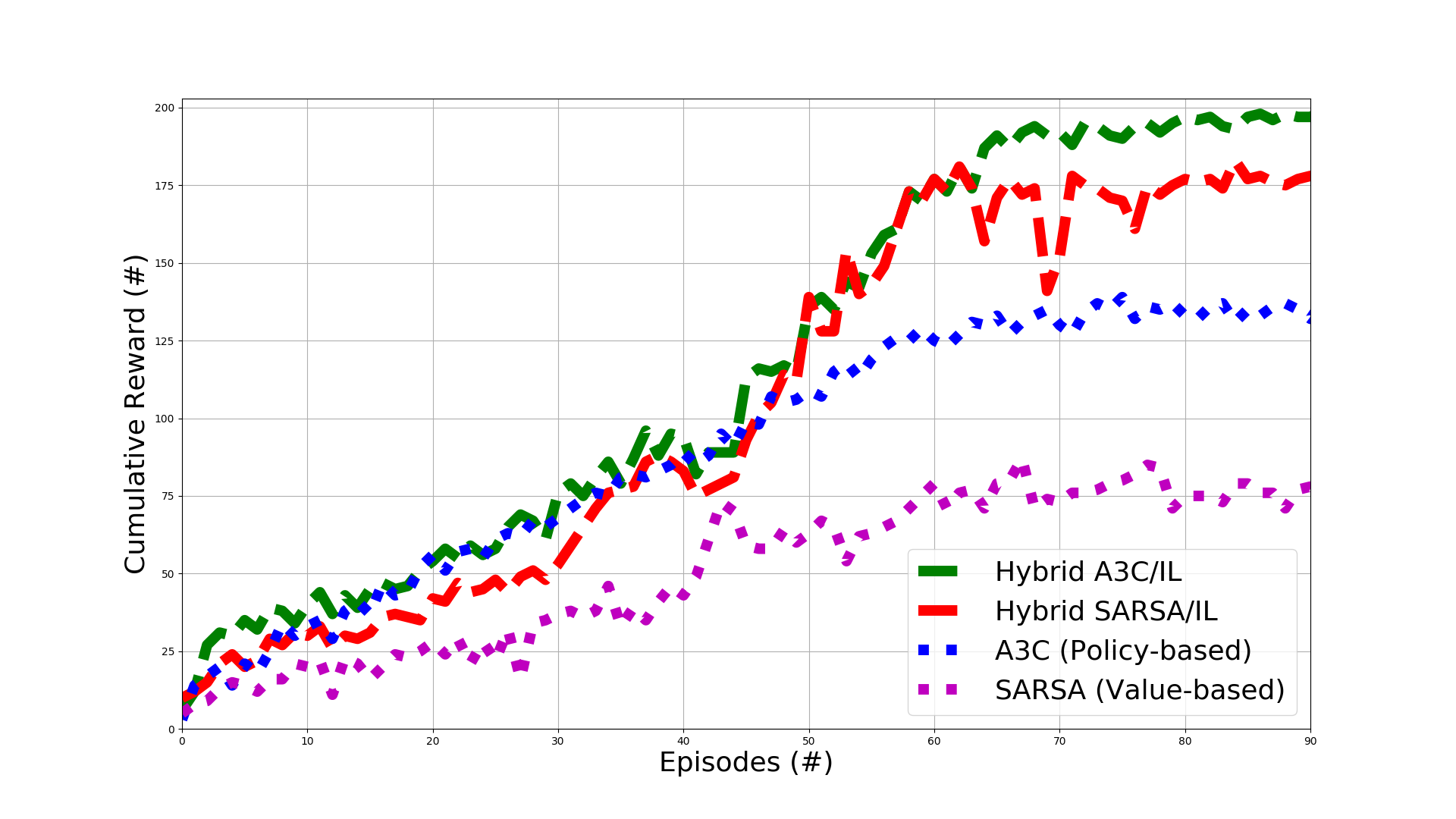}
         \caption{$ $}
        \label{fig:res1}
     \end{subfigure}
     \begin{subfigure}[b]{0.49\textwidth}
         \centering
         \includegraphics[width=\textwidth]{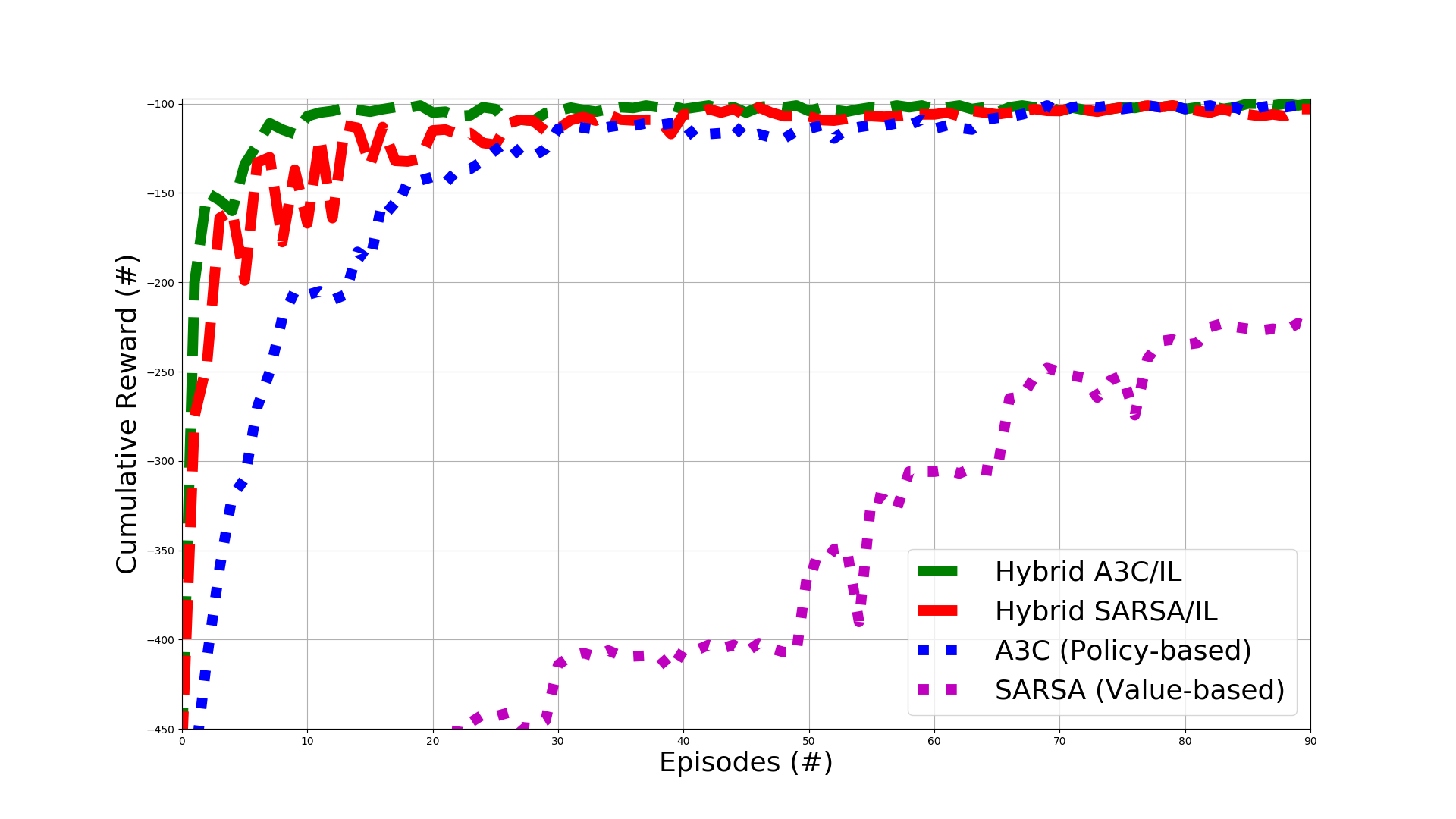}
         \caption{$ $}
        \label{fig:res 2}
     \end{subfigure}
 \centering
 \caption{Experimental results:(A) Continuous classic cart-pole  Openai Gym environment,(B) continuous classic mountain-car  Openai-Gym environment}
\end{figure}

\paragraph{}
The results of applying different algorithms (See Algorithms \ref{Algorithm SARSA}, \ref{Algorithm A3C}, \ref{Algorithm Hybrid SARSAIL}, and \ref{Algorithm HybridA3CIL}) on continuous classic cart-pole in OpenAI-Gym environment is presented in Fig 5(A). Each step, the agent is rewarded to balance with the horizontal axes. The results for both “Hybrid A3C/IL” and “Hybrid SARSA/IL” show that the proposed algorithms based on integration imitation learning and reinforcement learning can overcome the stand-alone reinforcement learning, A3C and SARSA. Fig \ref{fig:res1} presents Hybrid A3C/IL converges faster (in Episode \# 70) than Hybrid SARSA/IL. The reason of accelerating the convergence by Hybrid A3C/IL generally is based on the accuracy of the policy-based reinforcement learning in continuous environments. This is proven by even comparing standalone SARSA and A3C presented by blue and pink dots in the figure. It shows that the value-based reinforcement learning (SARSA here) in continuous environment like cart-pole is not satisfying regarding to data efficiency. Finally Hybrid A3C/IL increases data efficiency of the cart-pole environment by 53.8\% and 14.2\% compared to SARSA, A3C and Hybrid SARSA/IL, respectively.
\paragraph{}
Also, the achievements of utilizing Algorithms \ref{Algorithm SARSA}, \ref{Algorithm A3C}, \ref{Algorithm Hybrid SARSAIL}, and \ref{Algorithm HybridA3CIL} on continuous classic Mountain Car in OpenAI-Gym environment is shown in Fig \ref{fig:res 2}. In this environment, the agent receives punishment (negative reward) for each step of an episode. The maximum performance of this environment is -100 cumulative rewards and it shows that the minimum number of steps in an epoch to gain the flag on top of the hill in the continuous environment is 100. Like Cart-pole environment, the results for both “Hybrid A3C/IL” and “Hybrid SARSA/IL” show that the proposed algorithms based on the integration imitation learning and reinforcement learning outperform the A3C and SARSA, as examples of policy-based and value-based RL. Fig 5(B) presents Hybrid A3C/IL and Hybrid SARSA/IL converges faster the two RL algorithms. However, Hybrid SARSA/IL shows lots of oscillations before stabling at episode \# 20. The reason for these fluctuations is that value-iteration based RL cannot act well in complicated continuous environment regarding exploration and evaluation the tuple of state and action. Hybrid A3C/IL and Hybrid SARSA/IL, both increase data efficiency about 60\% and 33.4\% compared to SARSA and A3C, respectively.
\section{Conclusion}
In this paper, a novel approach is proposed which combines IL with different types of RL methods, namely State–action–reward–state–action (SARSA) and Asynchronous Advantage Actor-Critic Agents (A3C), to take advantage of both IL and RL methods. Also, we address how to effectively leverage the teacher’s feedback for the agent learner to learn sequential decision-making policies. The results of this study on simple OpenAI-Gym environment shows that Hybrid A3C/IL increases data efficiency of the cart-pole environment by 85.7\%, 53.8\% and 14.2\% compared to SARSA, A3C and Hybrid SARSA/IL, respectively. Also, the results on a complicated OpenAI-Gym environment present that Hybrid A3C/IL and Hybrid SARSA/IL, both increase data efficiency about 60\% and 33.4\% compared to SARSA and A3C, respectively.

\bibliographystyle{unsrt}  
\bibliography{references}  

\end{document}